\documentclass[accepted]{article}

\usepackage{aistats20221}

\usepackage[table]{xcolor}
\usepackage{bm}
\usepackage{amsmath,amsfonts,amsthm, amssymb}
\usepackage{mathrsfs}  
\usepackage{array}
\usepackage{xcolor, soul}
\usepackage{graphicx}
\usepackage{enumerate}
\usepackage{widetext}
\usepackage{multirow}
\usepackage{wrapfig}
\usepackage{bbm}
\usepackage{color}
\usepackage{times}
\usepackage{subcaption}
 \usepackage{verbatim}
\usepackage[utf8]{inputenc} %
\usepackage{enumitem}
\setitemize{noitemsep,topsep=0pt,parsep=0pt,partopsep=0pt}
\usepackage[belowskip=-4pt,aboveskip=0pt]{caption}

\usepackage{hhline}
\usepackage{float} 
\usepackage{booktabs}       %
\usepackage{xcolor,colortbl}

\definecolor{Gray}{gray}{0.85}
\definecolor{celestialblue}{rgb}{0.29, 0.65, 0.82}

\newcommand{\bl[1]}{{\bf #1}}

\newcommand{\Qset}{\mathcal{Q}}

\newcommand{\id}{\mathbb{I}}

\newtheorem{theorem}{Theorem}
\newtheorem{lemma}[theorem]{Lemma}

\newtheorem{definition}[theorem]{Definition}

\begin{document}

\twocolumn[

\aistatstitle{Choice functions based multi-objective Bayesian optimisation}

\aistatsauthor{ Alessio Benavoli \And Dario Azzimonti \And  Dario Piga }

\aistatsaddress{School of Computer Science and Statistics,\\ Trinity College Dublin, Ireland \And  Dalle Molle Institute for Artificial\\  Intelligence Research (IDSIA)\\ USI/SUPSI, Manno, Switzerland \And Dalle Molle Institute for Artificial\\ Intelligence  Research (IDSIA)\\  USI/SUPSI, Manno, Switzerland } ]

\begin{abstract}
In this work we introduce a new framework for multi-objective Bayesian optimisation where  the multi-objective functions can only be accessed via choice judgements, such as ``I pick options ${\bf x}_1,{\bf x}_2,{\bf x}_3$ among this set of five options ${\bf x}_1,{\bf x}_2,\dots,{\bf x}_5$''. The fact that the option ${\bf x}_4$ is rejected means that there is at least one option  among the selected ones ${\bf x}_1,{\bf x}_2,{\bf x}_3$ that I strictly prefer over ${\bf x}_4$ (but I do not have to specify which one). We assume that there is a latent vector function ${\bf f}$ for some dimension $n_e$ which embeds the options into the real vector space of dimension $n_e$, so that the choice set can be represented through a Pareto set of non-dominated options. By placing a Gaussian process prior on  ${\bf f}$ and deriving a novel likelihood model for choice data, we propose a Bayesian framework for choice functions learning. We then apply this surrogate model to solve a novel multi-objective Bayesian optimisation from choice data problem.
\end{abstract}

\section{Introduction}
\label{sec:intro}

In many practical situations a user is able to select the best options among a finite set of choices, however they are unable to state explicitly the motivations for their choices. A notable example is in industrial applications where the manufactured product has to satisfy several qualitative requirements that are known to trained staff, but such requirements were never expressed explicitly. In such cases, the definition of quantitative objectives would allow for an explicit multi-objective optimization which would lead to better options. However, measuring the objectives in a quantitative way is often technically difficult and costly. In this context, we would like to 
improve the quality of the manufactured product using directly  the feedback provided by the user's choices (``these products are better than those''), i.e. we would like to learn the ``choice function'' of the user and find the inputs that optimize this function. In this paper we propose a Bayesian framework to learn choice functions from a dataset of observed choices. Our framework learns a latent mapping of objectives that are consistent with the given choices, therefore we are also able to optimize them with a multi-objective Bayesian optimization algorithm. 

\section{Background}
The main contributions of this paper leverage four topics: (1) Bayesian Optimisation (BO); (2) preferential BO; (3) multi-objective BO; (4) choice functions learning. In this section we briefly review the state of the art of each topic.

\subsection{Bayesian Optimisation (BO)}
BO \cite{jones1998efficient} aims to find the global maximum of an unknown function which is expensive to evaluate. 
For a scalar real-valued function $g$ on a domain $\Omega \subset \mathbb{R}^{n_x}$, the goal is to find a global  maximiser   
$
{\bf x}^o =\arg \max_{{\bf x} \in \Omega} g({\bf x})
$. BO formulates this  as  a   sequential  decision  problem -- a trade-off between learning about the underlying function $g$  (exploration) and capitalizing on this information in order to find the optimum ${\bf x}^o$ (exploitation).
BO relies on a probabilistic surrogate model, usually a Gaussian Process (GP) \cite{rasmussen2006gaussian}, to provide a posterior distribution over $g$ given a dataset $\mathcal{D}=\{({\bf x}_i, g({\bf x}_i)): ~~i=1,2,\dots,N\}$ of previous evaluations of $g$. It then employs an \textit{acquisition function} (e.g. Expected Improvement \cite{jones1998efficient,mockus1978application}, Upper Credible Bound \cite{srinivas2009gaussian})  to select the next  candidate option (solution) ${\bf x}_{N+1}$. While the true function $g$ is  expensive-to-evaluate, the surrogate-based acquisition function is not, and it can thus be efficiently optimized to compute an optimal candidate  to be evaluated on $g$. This process is repeated sequentially until some stopping criterion is achieved.       

\subsection{Preferential Bayesian Optimisation (PBO)}
In many applications,  evaluating $g$ can be either too costly or not always possible. In these cases, the  objective function $g$ may only be accessed via preference judgments, such as ``this is better than that'' between two candidate options ${\bf x}_i,{\bf x}_j$ like in A/B tests or recommender systems (pairwise comparisons are usually called \textit{duels} in the BO and bandits literature).
In such situations,  PBO \cite{shahriari2015taking}  can be used. This approach requires the agent to  simply  compare the final outcomes of two different candidate options and indicate which they prefer, that is the evaluation is binary either ${\bf x}_i$ ``better than'' ${\bf x}_j$ or ${\bf x}_i$ ``worse than'' ${\bf x}_j$.  %

In the PBO context, the state-of-the-art surrogate model is based on a method for preference learning developed in \cite{ChuGhahramani_preference2005}. This method assumes that there is an unobservable latent function value $f({\bf x}_i)$ associated with
each training sample ${\bf x}_i$, and that the function values $\{f({\bf x}_i): ~~i=1,2,\dots,N\}$ preserve the preference relations observed in
the dataset, that is  $f({\bf x}_i)\geq f({\bf x}_j)$ whenever ${\bf x}_i$ ``better than'' ${\bf x}_j$. %
As in the BO setting, by starting with a GP prior on $f$ and, by using the likelihood defined in \cite{ChuGhahramani_preference2005}, we obtain a posterior distribution over $f$ given a dataset of preferences. This posterior distribution is not a GP and several approximations \cite{ChuGhahramani_preference2005,houlsby2011bayesian} were proposed. In \cite{houlsby2011bayesian}, the authors showed that GP preference learning is equivalent to GP classification with a transformed  kernel function. By using this reformulation, the authors easily derive two approximations for the posterior based on (i) the Laplace Approximation (LP) \cite{mackay1996bayesian,williams1998bayesian}; (ii) Expectation Propagation (EP) \cite{minka2001family}. The LP approximation was then used to develop a framework for PBO \cite{shahriari2015taking} and a new acquisition function, inspired by Thomson sampling, was proposed in \cite{gonzalez2017preferential}. More recently, \cite{benavoli2020preferential} showed that the posterior of GP preference learning is a Skew GP \cite{Benavoli_etal2020,benavoli2021}. Based on this exact model, the authors derived a PBO framework  which outperformed both LP and EP based PBO.     
Although in this work we focus on GPs as surrogate model, it is worth to mention alternative approaches for PBO developed by \cite{BDG08,pmlr-v32-zoghi14,sadigh2017active,bemporad2019active}.

\begin{figure}[htp]
  \begin{center}
    \includegraphics[width=5cm]{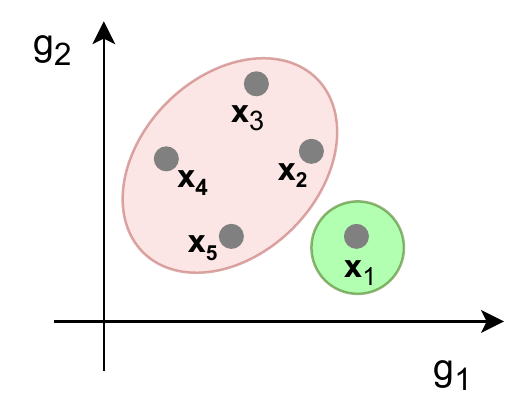}
  \end{center}
  \caption{Batch PBO: the preferred option is in green.}   
  \label{fig:1}
\end{figure}

PBO was recently extended \cite{siivola2020preferential} to the batch case by allowing agents to express  preferences for a batch of options. However, as depicted in Figure \ref{fig:1} where an agent expresses preferences among 5 options, 
in batch PBO there can be only one batch winner. In fact, PBO  assumes that two options are always comparable.\footnote{More precisely, the underlying GP-based model implies a total order and so two options may also be equivalent. When PBO is applied to the multi-objective case such as for instance $[g_1(x_i),g_2(x_i)]$, it is therefore assumed that the agent's preferences are determined by a weighted combination of the objectives $w_1 g_1(x_1)+w_2g_2(x_2)$.}  

\subsection{Multi-objective (MO) optimization} %
The goal of MO optimization is to identify the set of 
\textit{Pareto optimal} options (solutions) such that any improvement in one objective means deteriorating another. Without loss of generality, we assume the goal is to maximize all objectives.
Let ${\bf g}({\bf x}):\Omega \rightarrow \mathbb{R}^{n_o}$ be a vector-value objective function with ${\bf g}({\bf x})=[g_1({\bf x}),\dots,g_{n_o}({\bf x})]^\top$, where $n_0$ is the number of objectives. We recall the notions of Pareto dominated options and non-dominated set.      
 
\begin{definition}[Pareto dominate option]
\label{def:paretodom}
 Consider a set of options $\mathcal{X} \subset \Omega$. An option ${\bf x}_1 \in \mathcal{X}$ is  said to Pareto dominate   another option ${\bf x}_2 \in \mathcal{X}$, denoted as  ${\bf x}_1 \succ  {\bf x}_2$, if both the following conditions are true: 
 \begin{enumerate}
  \item for all $j \in \{1,2,\dots,n_o\}$, $g_j ({\bf x}_1) \geq  
g_j ({\bf x}_2 )$;
  \item  $\exists ~j \in \{1,2,\dots,n_o\}$, such that $g_j ({\bf x}_1) >
g_j ({\bf x}_2 )$.  
 \end{enumerate}
\end{definition}

\begin{definition}[Non-dominated set]
\label{def:paretoset}
Among a set of options  $A=\{{\bf x}_1,\dots,{\bf x}_m\}$, the non-dominated set of options  $A'$ are those that are not  dominated by any member of  $A$, i.e.
\begin{equation*}
A' = \{ {\bf x} \in A : \nexists {\bf x}^\prime \in A \text{ such that } {\bf x}^\prime \succ {\bf x} \}.
\end{equation*}
\end{definition}
Given the set of options  $\mathcal{X}$, MO aims to find the non-dominated set of options  $\mathcal{X}^{nd}$, called the \textit{Pareto set}. The set of evaluations $\mathbf{g}(\mathcal{X}^{nd})$ is called \textit{Pareto front}.    
     
MO BO have only be developed for standard (non-preferential) BO, where  multi-objectives can directly be evaluated. Many approaches rely on scalarisation to transform the MO problem into a single-objective one, like ParEGO \cite{knowles2006parego} and TS-TCH \cite{paria2020flexible}
(which randomly scalarize the objectives and use Expected Improvement and, respectively, Thompson Sampling). 
\cite{keane2006statistical}  derived an \textit{expected improvement} criterion with respect to multiple objectives. \cite{ponweiser2008multiobjective} proposed an \textit{hypervolume-based infill} criterion, where the improvements are measured in terms of hypervolume (of the Pareto front) increase. Other acquisition functions have been proposed in \cite{emmerich2011hypervolume,picheny2015multiobjective,hernandez2016predictive,wada2019bayesian,belakaria2019max}.
The most used acquisition function for MO BO is \textit{expected hypervolume improvement}. In fact, maximizing the hypervolume has been shown to produce very accurate estimates \cite{zitzler2003performance,couckuyt2014fast,hupkens2015faster,emmerich2016multicriteria,yang2017computing,yang2019multi} of the Pareto front. 

\subsection{Choice function}
\label{sec:choice}
Individuals are often confronted with the situation of choosing between several options (alternatives). These alternatives can be goods that are going to be purchased, candidates in elections, food etc. 

\begin{figure}[htp]
  \begin{center}
    \includegraphics[width=5cm]{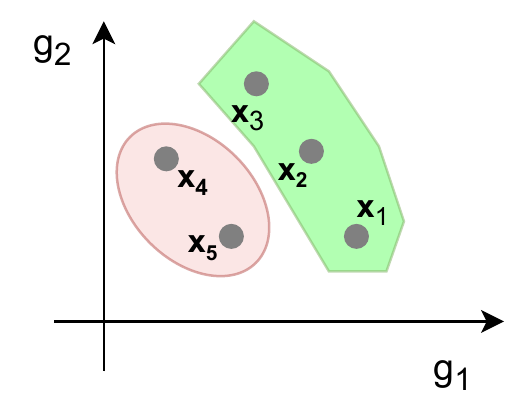}
  \end{center}
  \caption{Example of choice function for $A=\{x_1,x_2,\dots,x_5\}$: $C(A)=\{x_1,x_2,x_3\}$ highlighted in green and $R(A)=\{x_4,x_5\}$ in red.}   
  \label{fig:2}
\end{figure}

We model options, that an agent has to choose, as real-valued vectors  ${\bf x} \in \mathbb{R}^{n_x}$ and identify the sets of options as finite subsets of $\mathbb{R}^{n_x}$. Let  $\Qset$ denote the set of all such finite subsets of $\mathbb{R}^{n_x}$. 
\begin{definition}
 A choice function $C$ is a set-valued operator on sets of options. More precisely, it is a map $C: \Qset \rightarrow \Qset$ such that, for any set of options $A \in \Qset$, the corresponding value of $C$ is a subset $C(A)$ of $A$ (see for instance \cite{aleskerov2007utility}). 
\end{definition}
The interpretation of choice function is as follows. For a given option set $A \in \Qset$, the statement that an option ${\bf x}_j \in A$ is rejected from $A$ (that is, ${\bf x}_j  \notin C(A)$) means that there is at least one option  ${\bf x}_i \in A$ that an agent strictly prefers over ${\bf x}_j$. The set of rejected options is denoted by $R(A)$ and is equal to $A\backslash C(A)$. Therefore choice functions represent non-binary choice models, so they are more general than preferences. \\
It is important to stress again that the statement ${\bf x}_j  \notin C(A)$ implies there is at least one option  ${\bf x}_i \in A$ that an agent strictly prefers over ${\bf x}_j$. However, the agent is not required to tell us which option(s) in $C(A)$ they strictly prefer to ${\bf x}_j$. This makes choice functions a very easy-to-use tool to express choices. As depicted in Figure \ref{fig:2}, the agent needs to tell us only the options they selected (in green) without providing any justification for their choices (we do not know which option in the green set dominates ${\bf x}_4$).

By following this interpretation, the set $C(A)$ can also be seen as the \textit{non-dominated set} in the Pareto sense for some latent function. In other words, let us assume that there is a latent vector function ${\bf g}({\bf x}_i)=[g_1({\bf x}_i),\dots,g_{n_e}({\bf x}_i)]^\top$, for some dimension $n_e$, which embeds the options ${\bf x}_i$ into a  space $\mathbb{R}^{n_e}$. The choice set can then be represented through a Pareto set of non-dominated options. For example, in Fig.~\ref{fig:2}, $n_e=2$. This approach was proposed in \cite{pfannschmidt2020learning} to learn choice functions.  In particular, to learn the latent vector function, the authors devise a differentiable loss function based on a hinge loss. Furthermore, they add two additional terms to the loss function: (i) an $L^2$ regularization term; (ii) a multidimensional scaling (MDS) loss to ensure that options close to each other in the inputs space $\mathcal{X}$ will also be close in the embedding space $\mathbb{R}^{n_e}$. This loss function is then used to learn a (deep) multi-layer perceptron to represent the embedding.

\subsection{Contributions}
In this work, we devise a novel multi-objective PBO based on choice functions. We follow the interpretation of choice functions as set function that select non-dominated sets for an unknown latent function. First we derive a Bayesian framework to learn the function from a dataset of observed choices. This framework is based on a Gaussian Process prior on the unknown latent function vector.\footnote{Compared to the approach proposed in \cite{pfannschmidt2020learning}, the GP-based model is more sound -- no multidimensional scaling is necessary -- and it is a generative model.} We then build an acquisition function to select the best next options to evaluate. We compare this method against an oracle that knows the true value of the latent functions and we show that, by only working with choice function evaluations, we converge to the same results.

\section{Bayesian learning of Choice functions}
In this work we consider options ${\bf x}\in \mathbb{R}^{n_x}$ and, for ${\bf x}\in \mathbb{R}^{n_x}$, we model each  latent function in the vector ${\bf f}({\bf x})=[f_1({\bf x}),\dots,f_{n_e}({\bf x})]^\top$ as an independent GP \cite{rasmussen2006gaussian}:
\begin{equation}
\label{eq:prior}
 f_j({\bf x}) \sim \text{GP}_j(0,k_j({\bf x},{\bf x}')), ~~~~j=1,2,\dots,n_e.
\end{equation}
 Each GP is fully specified by its kernel function $k_j(\cdot,\cdot)$, which specifies the covariance of the latent function between any two points. In all experiments in this paper, the GP kernel is Matern 3/2 \cite{rasmussen2006gaussian}.

\subsection{Likelihood for general Choice functions}
Having defined the prior on ${\bf f}$, we can now focus on the likelihood.
We propose a new likelihood to model the observed choices of the agent. Given a set of observed choices $\mathcal{D}=\{(A_k,C(A_k)): \text{ for } k=1,\dots,N\}$, we are interested in learning a Pareto-embedding ${\bf f}$ coherent with this data in the sense that $C(A_k)=P_{{\bf f}}(A_k)$, where $P_{{\bf f}}(A_k)$ denotes the Pareto non-dominated options in $A_k$. 

Assume that $A_k=\{{\bf x}_1,\dots,{\bf x}_m\}$ and let $I \subset \{1,2,\dots,m\}$ be the subset of indices of the options in $C(A_k)$, let $J_k$ be equal to $\{1,2,\dots,m\}\backslash I_k$ and let $D=\{1,2,\dots,n_e\}$ the vector of dimensions of the latent space. Based on Definition \ref{def:paretodom}, the choice of the agent expressed via $C(A_k)$ implies that:
\begin{align}
 \label{eq:likcond1}
 \neg &\left( \min_{d \in D} (f_d({\bf x}_i)-f_d({\bf x}_j))< 0, ~~\forall i \in I_k\right), \forall j \in J_k,\\
  \label{eq:likcond2}
 &\min_{d \in D} (f_d({\bf x}_p)-f_d({\bf x}_i))< 0, ~~\forall i, p \in I_k, p\neq i.
\end{align}
These equation express the conditions in Definition~\ref{def:paretodom}. Condition \eqref{eq:likcond1} means that, for each option $x_j \in J_k$, it's not true ($\neg$ stands for logical negation) that all options in $I_k$ are worse than $x_j$, i.e. there is at least an option in $I_k$ which is better than $x_j$. Condition   \eqref{eq:likcond2} means that, for each option in $I_k$, there is no better option in $I_k$. This requires that the latent functions values of the options should be consistent with the choice function implied relations. Given $A_k,C(A_k)$, the likelihood function $p(C(A_k),A_k|{\bf f})$ is one when \eqref{eq:likcond1}-\eqref{eq:likcond2} hold and zero otherwise.

In practice not all choices might be coherent and we can treat this case by considering that 
the latent vector function ${\bf f}({\bf x}_i)=[f_1({\bf x}_i),\dots,f_{n_e}({\bf x}_i)]^\top$ is corrupted by a Gaussian  noise ${\bf v}_i$ with zero mean vector and  covariance $\sigma^2 \id_{n_e}$.\footnote{We assume the noise variance is the same in each dimension but this  can  easily be relaxed.}
Then we require  conditions \eqref{eq:likcond1} and  \eqref{eq:likcond2} to only hold probabilistically. This leads to the following likelihood function for the pair $A_k,C(A_k)$:

\begin{widetext}
	\small
\begin{align}
\nonumber
&p(C(A_k),A_k|{\bf f})=
\prod_{j \in J_k}\Bigg(1-\int \prod_{i \in I_k}\left(\mathcal{I}_{(-\infty,0)}\left(\min_{d \in D} (f_d({\bf x}_i)+v_{di}-f_d({\bf x}_j)-v_{dj})\right)N({\bf v}_{i};0,\sigma^2\id_d)d{\bf v}_{i} \right)N({\bf v}_{j};0,\sigma^2\id_d)d{\bf v}_{j}\Bigg)\\
  \label{eq:like0}
 & \prod_{i, p \in I_k, p \neq i}\int \left(1-\mathcal{I}_{[0,\infty)}\left( \min_{d \in D} (f_d({\bf x}_p)+v_{dp}-f_d({\bf x}_i)-v_{di})\right)\right)N({\bf v}_{p};0,\sigma^2\id_d)N({\bf v}_{i};0,\sigma^2\id_d)d{\bf v}_{p}d{\bf v}_{i},
  \end{align}
  \end{widetext}
   where $v_{di}$ denotes the $d$-th component of the vector ${\bf v}_i$ and $\mathcal{I}_B$ is the indicator function of the set $B$. 
We now provide two results which allows us to simplify \eqref{eq:like0}. We first compute the integral in the third product in \eqref{eq:like0}.
\begin{lemma}
\label{lem:0}
 \begin{equation}
 \begin{aligned}
   &\int \mathcal{I}_{[0,\infty)}\left(\min_{d \in D} (f_d({\bf x}_p)+v_{dp}-f_d({\bf x}_j)-v_{dj})\right)\\
   &N({\bf v}_{p};0,\sigma^2\id_d)N({\bf v}_{j};0,\sigma^2\id_d)d{\bf v}_{p}d{\bf v}_{j}\\
   &=\prod_{d \in D} \Phi\left(\frac{f_d({\bf x}_p)-f_d({\bf x}_j)}{\sqrt{2}\sigma}\right).
   \end{aligned}
 \end{equation}
\end{lemma}
All proofs are in the supplementary material.
We now focus on the first integral in \eqref{eq:like0}, which can be simplified as follows.
\begin{lemma}
\label{lem:2}
 \begin{equation}
 \label{eq:lemma2}
 \begin{aligned}
 &\int \prod_{i \in I_k} \Big(\mathcal{I}_{(-\infty,0)}\left(\min_{d \in D} (f_d({\bf x}_i)+v_{di}-f_d({\bf x}_j)-v_{dj})\right)\\
 &N({\bf v}_{i};0,\sigma^2\id_d)d{\bf v}_{i} \Big)N({\bf v}_{j};0,\sigma^2\id_d)d{\bf v}_{_j}\\
   &=\int \prod_{i \in I_k} \left[1- \prod_{d \in D} \Phi\left(\frac{f_d({\bf x}_i)-f_d({\bf x}_j)-v_{dj}}{\sigma}\right)\right] \\
   &N({\bf v}_{j};0,\sigma^2\id_d)d{\bf v}_{j}.
   \end{aligned}
 \end{equation}
\end{lemma}  
Note that eq.~\eqref{eq:lemma2} is an expectation (with respect to $ N({\bf v}_{j};0,\sigma^2\id_d)$) of a product of Gaussian CDFs $\Phi(\cdot)$ whose argument only depends on $v_{dj}$. We can thus write the above multidimensional integral as a sum of products of univariate integrals which can be computed efficiently, for instance by using a Gaussian quadrature rule.

Therefore, the likelihood of the choices $\mathcal{D}=\{(A_k,C(A_k)): \text{ for } k=1,\dots,N\}$ given the latent vector function ${\bf f}$ can then be written as follows.
\begin{theorem}
	\label{thm:1}
The likelihood is
 \begin{equation}
  p(\mathcal{D}|{\bf f})=\prod_{k=1}^N p(C(A_k),A_k|{\bf f})
 \end{equation}
with
  \begin{equation}
  \label{eq:likelihood}
 \begin{aligned}
      &p(C(A_k),A_k|{\bf f})\\
      =&\prod_{j \in J_k}\Bigg(1-\int \prod_{i \in I_k} \left(1- \prod_{d \in D} \Phi\left(\frac{f_d({\bf x}_i)-f_d({\bf x}_j)-v_{dj}}{\sigma}\right)\right)\\
      &N({\bf v}_{j};0,\sigma^2\id_d)d{\bf v}_{j}\Bigg)\\
   & \prod_{i, p \in I_k, p\neq i}\left( 1-\prod_{d \in D} \Phi\left(\frac{f_d({\bf x}_p)-f_d({\bf x}_j)}{\sqrt{2}\sigma}\right)\right).\\
   \end{aligned}
 \end{equation}
\end{theorem}

\subsubsection{Likelihood for batch preference learning}
In case $n_e=1$ (the latent dimension is one), we have that $|C(A_k)|=1$. This means the agent always selects a single best option. In this case, the likelihood \eqref{eq:likelihood} simplifies to
  \begin{equation}
  \label{eq:likelihood1}
 \begin{aligned}
      p(C(A_k),A_k|{\bf f})
      =&\int \prod_{j \in J_k}\Phi\left(\frac{f({\bf x}_i)-f({\bf x}_j)-v_{j}}{\sigma}\right)\\
      &N(v_{j};0,\sigma^2)dv_{j}.\\
   \end{aligned}
 \end{equation}
The above likelihood is equal to the batch likelihood derived in \cite[Eq.3]{siivola2020preferential} and reduces to the likelihood derived in \cite{ChuGhahramani_preference2005} when $|R(A_k)|=1$ (that is the batch 2 case, i.e., $|A_k|=2$). This shows that the likelihood in \eqref{eq:likelihood} encompasses batch preference-based models.

\subsection{Posterior}
\label{sec:posterior}
The posterior probability of ${\bf f}$ is
\begin{equation}
\label{eq:post}
p({\bf f}|\mathcal{D})=\frac{p({\bf f})}{p(\mathcal{D})} \prod_{k=1}^N p(C(A_k),A_k|{\bf f}),
\end{equation}
where the prior over the component of ${\bf f}$ is defined in \eqref{eq:prior}, the likelihood is defined in \eqref{eq:likelihood}
and the probability of the evidence is $p(\mathcal{D})= \int p(\mathcal{D}|{\bf f})p({\bf f}) d{\bf f}$. The posterior $p({\bf f}|\mathcal{D})$ is intractable because it is neither Gaussian nor Skew Gaussian Process (SkewGP) distributed. 
In this paper we propose an approximation schema for the posterior similar to the one proposed in \cite{benavoli2020preferential,benavoli2021}. In \cite{benavoli2020preferential}, an analytical formulation of the posterior is available, the marginal likelihood is approximated with a lower bound and inferences are computed with an efficient rejection-free slice sampler \cite{gessner2019integrals}. In~\cite{benavoli2020preferential,benavoli2021} such approximation schema showed better performance in active learning and BO tasks than LP and EP. Here we do not have an analytical formulation for the posterior therefore we use a Variational (ADVI) approximation \cite{kucukelbir2015automatic} of the posterior to learn the hyperparameters $\boldsymbol{\theta}$ of the kernel and, then, for fixed hyperparameters, we compute the posterior of  $p({\bf f}|\mathcal{D},\boldsymbol{\theta})$ via elliptical slice sampling (ess) \cite{pmlrv9murray10a}.\footnote{We implemented our model in PyMC3   \cite{salvatier2016probabilistic}, which provides implementations of ADVI and ess. Details about number of iterations and tuning are reported in the supplementary.} 

\subsection{Prediction and Inferences}
\label{sec:prediction}
Let $A^*=\{{\bf x}^*_1,\dots,{\bf x}^*_m\}$ be a set including $m$ test points and ${\bf f}^*=[{\bf f}({\bf x}_1^*),\dots,{\bf f}({\bf x}_m^*)]^\top$. The conditional predictive distribution $p({\bf f}^*|{\bf f})$ is Gaussian and, therefore, 
\begin{equation}
\label{eq:pred}
p({\bf f}^*|\mathcal{D})= \int p({\bf f}^*|{\bf f})p({\bf f}|\mathcal{D}) d{\bf f}
\end{equation}
 can be easily approximated as a sum using the samples from $p({\bf f}|\mathcal{D})$.
In choice function learning, we are interested in the inference: 
\begin{equation}
\label{eq:infer}
P(C(A^*),A^*|\mathcal{D})=\int p(C(A^*),A^*|{\bf f}^*) p({\bf f}^*|\mathcal{D}) d{\bf f}^*,
\end{equation}
which returns the posterior probability that the agent chooses the options $C(A^*)$ from the set of options $A^*$.
Given a finite set $\mathcal{X}$ (that is $A^*=\mathcal{X}$), we can use \eqref{eq:infer} to compute the probability that a subset of $\mathcal{X}$ is the set of non-dominated options. This provides an estimate of the Pareto-set  $\mathcal{X}^{nd}$ based on the learned GP-based latent model.

\section{Latent dimension selection}
\label{sec:lat}
In the previous section, we provided a Bayesian model for learning a choice function. The model $\mathcal{M}_{n_e}$ is conditional on the pre-defined latent dimension $n_e$ (that is, the dimension of the vector of the latent functions ${\bf f}({\bf x})=[f_1({\bf x}),\dots,f_{n_e}({\bf x})]^\top$). Although, it is sometimes reasonable to assume  the number of criteria defining the choice function to be known (and so the dimension $n_e$), it is crucial to develop a statistical method to select $n_e$. We propose a forward selection method. We start learning the model $\mathcal{M}_{1}$ and we increase the dimension $n_e$ in a stepwise manner (so learning $\mathcal{M}_{2},\mathcal{M}_{3},\dots$) until some model selection criterion is optimised. Criteria like AIC and BIC are inappropriate for the proposed GP-based choice  function model, since its nonparametric nature implies that the number of parameters increases also with the size of the data (as $n_e \times m$). We propose to use instead the \textit{Pareto Smoothed Importance sampling Leave-One-Out} cross-validation (PSIS-LOO) \cite{vehtari2017practical}. Exact cross-validation requires re-fitting the model with different training sets. Instead, PSIS-loo can be computed easily using the samples from the posterior. 

We define the Bayesian LOO estimate of out-of-sample predictive fit for the model in \eqref{eq:post}:
\begin{equation}
\label{eq:bloo}
 \varphi=\sum_{k=1}^N p(z_k|z_{-k}),
\end{equation}
where $z_k=(C(A_k),A_k)$, $z_{-k}=\{(C(A_i),A_i)\}_{i=1,i\neq k}^N$,
\begin{equation}
\label{eq:bloo1}
p(z_k|z_{-k})=\int p(z_k|{\bf f})p({\bf f}|z_{-k})d{\bf f}.
\end{equation}
As derived in \cite{gelfand1992model}, we can evaluate \eqref{eq:bloo1} using the samples from the full posterior, that is  ${\bf f}^{(s)}\sim p({\bf f}|\{z_k,z_{-k}\})=p({\bf f}|\mathcal{D})$ for $s=1,\dots,S$.\footnote{We compute these samples using elliptical slice sampling.} We first define the importance weights:
$$
w^{(s)}_k=\frac{1}{p(z_k|{\bf f}^{(s)})}\propto \frac{p({\bf f}^{(s)}|z_{-k})}{p({\bf f}^{(s)}|\{z_k,z_{-k}\})}
$$
and then approximate \eqref{eq:bloo1} as:
\begin{equation}
\label{eq:bloo2}
p(z_k|z_{-k})\approx \frac{\sum_{s=1}^S w^{(s)}_k p(z_k|{\bf f}^{(s)})}{\sum_{s=1}^S w^{(s)}_k }.
\end{equation}
It can be noticed that \eqref{eq:bloo2} is a function of $p(z_k|{\bf f}^{(s)})$ only, which can easily be computed from the posterior samples. Unfortunately,  a direct use of \eqref{eq:bloo2} induces instability because the importance
weights can have high variance. To address this issue, \cite{vehtari2017practical} applies a simple smoothing
procedure to the importance weights using a Pareto distribution (see \cite{vehtari2017practical} for details). In Section \ref{sec:latsim}, we will show that the proposed PSIS-LOO-based forward procedure works well in practice.

\section{Choice-based Bayesian Optimisation}
\label{sec:acq}
In the previous sections, we have introduced a GP-based model to learn latent choice functions from choice data. We will now focus on the acquisition component of Bayesian optimization.
In choice-based BO, we never observe the actual values of the functions. The data is $(\mathcal{X},\{C(A_k),A_k\}_{k=1}^N\})$, where 
$\mathcal{X}$ is the set of the $m$ training inputs (options), $A_k$ is a subset of $\mathcal{X}$ and $C(A_k) \subseteq A_k$ is the choice-set for the given options $A_k$. We denote the Pareto-set estimated using the GP-based model as $\hat{\mathcal{X}}^{nd}$.
In choice-based BO, the objective is to seek a new input point ${\bf x}$. Since ${\bf g}$ can only be queried via a choice function, this is obtained by optimizing w.r.t.\ ${\bf x}$ an acquisition function $\alpha({\bf x},\hat{\mathcal{X}}^{nd})$, where $\hat{\mathcal{X}}^{nd}$ is the current (estimated) Pareto-set. We define the acquisition function  $\alpha({\bf x},\hat{\mathcal{X}}^{nd})$, with the aim to find a point that dominates the points in $\hat{\mathcal{X}}^{nd}$. That is, given the set of options $A^*={\bf x} \cup \hat{\mathcal{X}}^{nd}$, we aim to find ${\bf x}$ such that $C(A^*)=\{{\bf x}\}$. 
 
 The acquisition function must also consider the trade-off between exploration and exploitation. Therefore, we propose an acquisition function   $\alpha({\bf x},\hat{\mathcal{X}}^{nd})$ which is equal to the $\gamma$\% (in the experiments we use $\gamma=95$) Upper Credible Bound (UCB) of $p(C(A^*),A^*|{\bf f}^*)$   with ${\bf f}^* \sim p({\bf f}^*|\mathcal{D})$, $A^*={\bf x} \cup \hat{\mathcal{X}}^{nd}$ and $C(A^*)=\{{\bf x}\}$. \\ 
 Note that the requirement for our acquisition function is strong. We could also define $\tilde{\alpha}({\bf x},\hat{\mathcal{X}}^{nd})$ with different objectives in mind. For example we could seek to find a point ${\bf x}$ which allows to reject at least one option in $\hat{\mathcal{X}}^{nd}$. We opted for UCB over the probability $p(C(A^*),A^*|{\bf f}^*)$ because it leads to a fast to evaluate acquisition function. In particular we only need to compute one probability for each new function evaluation. In future work we will study alternative approaches and the trade-off between more costly acquisition function evaluations and faster convergence. 
 
 After computing the maximum of the the acquisition function, denoted with ${\bf x}_{new}$, consistently with the definition of the acquisition function, we should query the agent to express their choice among the set of options in
 $A^*={\bf x}_{new} \cup \hat{\mathcal{X}}^{nd}$. However, $\hat{\mathcal{X}}^{nd}$ can be a very large set and human cognitive ability cannot efficiently compare more than five options. Therefore, by using the GP-based latent model, we select four options in $\hat{\mathcal{X}}^{nd}$ which have the highest probability of being dominated by ${\bf x}_{new}$ and query the agent on a five options set.\footnote{Details about the procedure we use to select these 4 options are reported in the supplementary material.}

\section{Numerical experiments}
First, we assume $n_e=n_o$ (that is we assume that the latent dimension is known) and evaluate the performance of our algorithm on (1) the tasks of learning choice functions; (2) the use of choice functions in multi-objective BO. 
Second, we evaluate the latent dimension selection procedure discussed in Section \ref{sec:lat} on simulated and real datasets.

\subsection{Choice functions learning}
\paragraph{Toy experiment}
We develop a simple toy experiment as a controlled
setting for our initial assessment. We consider the bi-dimensional vector function ${\bf g}(x)=[\cos(2x),-sin(2 x)]^\top$ with $x \in \mathbb{R}$.   

\begin{tabular}{c}\vspace{-0.3cm} 
~~~~~~~~~~~~~\includegraphics[height=3cm]{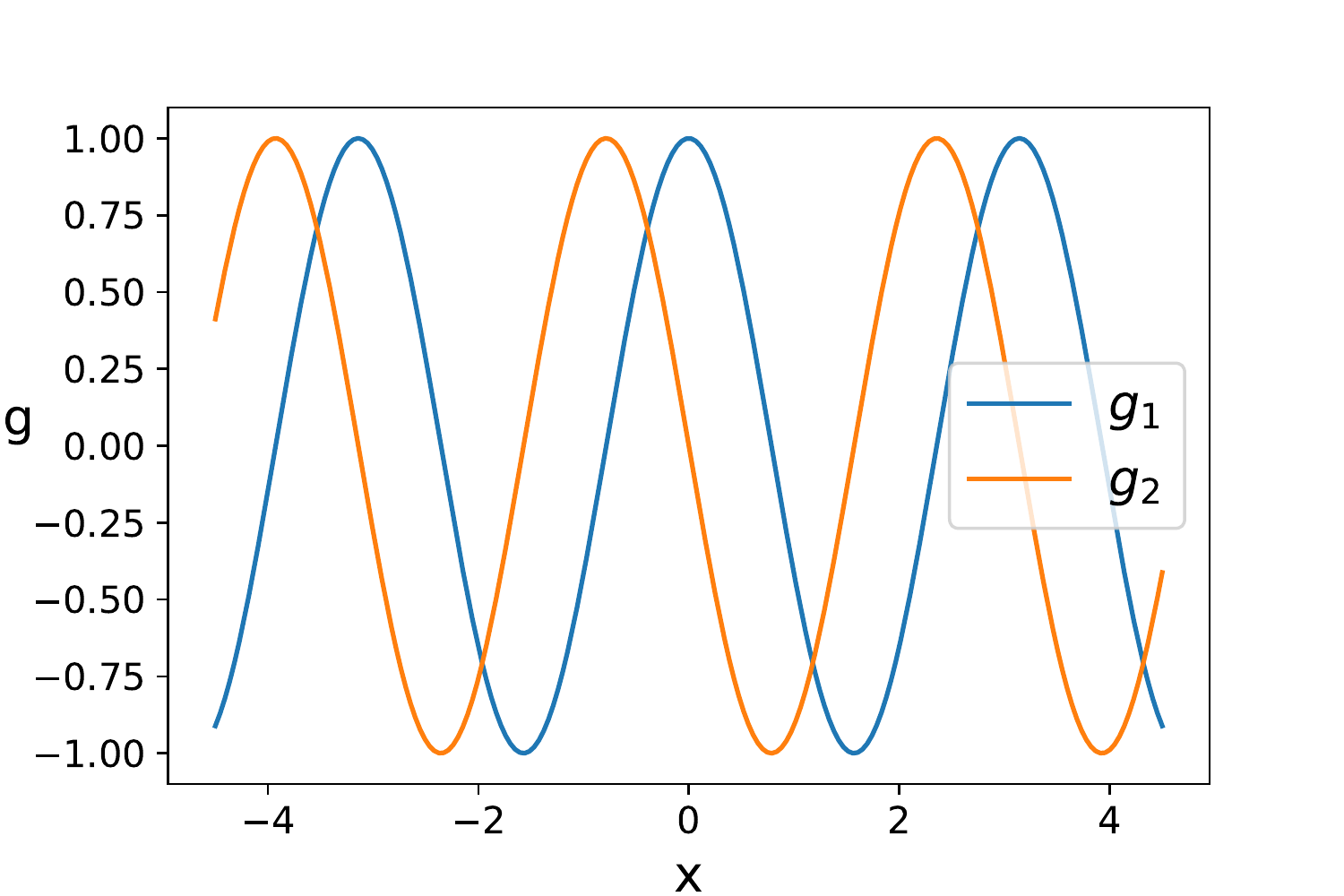}
\vspace{0cm}                                                                                                                                           \end{tabular}

\begin{figure*}
	\centering
	\begin{tabular}{cccc}
		\includegraphics[height=3cm,trim={0.8cm 0.0cm 1.5cm 0.0cm }, clip]{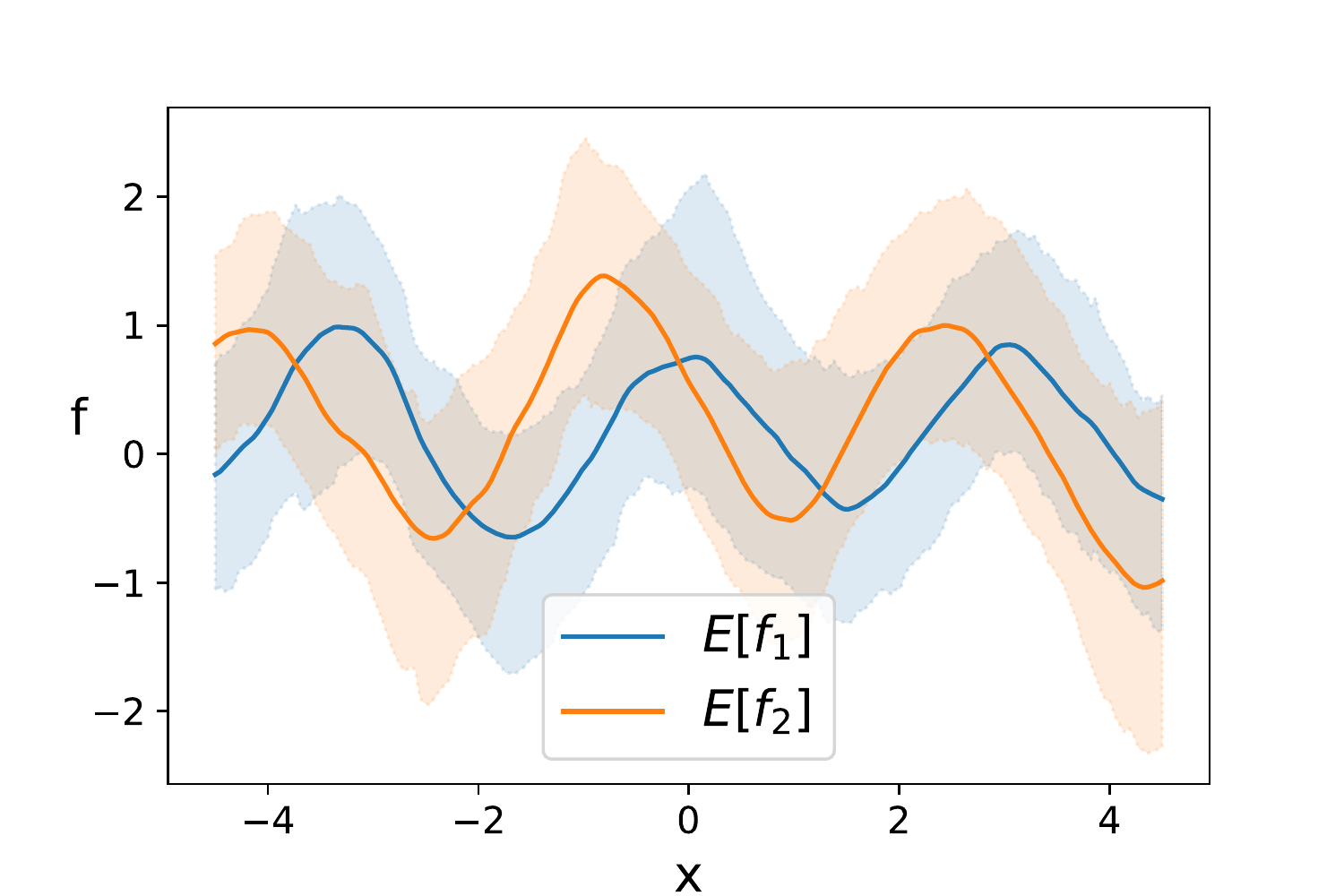} &
		\includegraphics[height=3.0cm,trim={0.8cm 0.0cm 1.5cm 0.0cm }, clip]{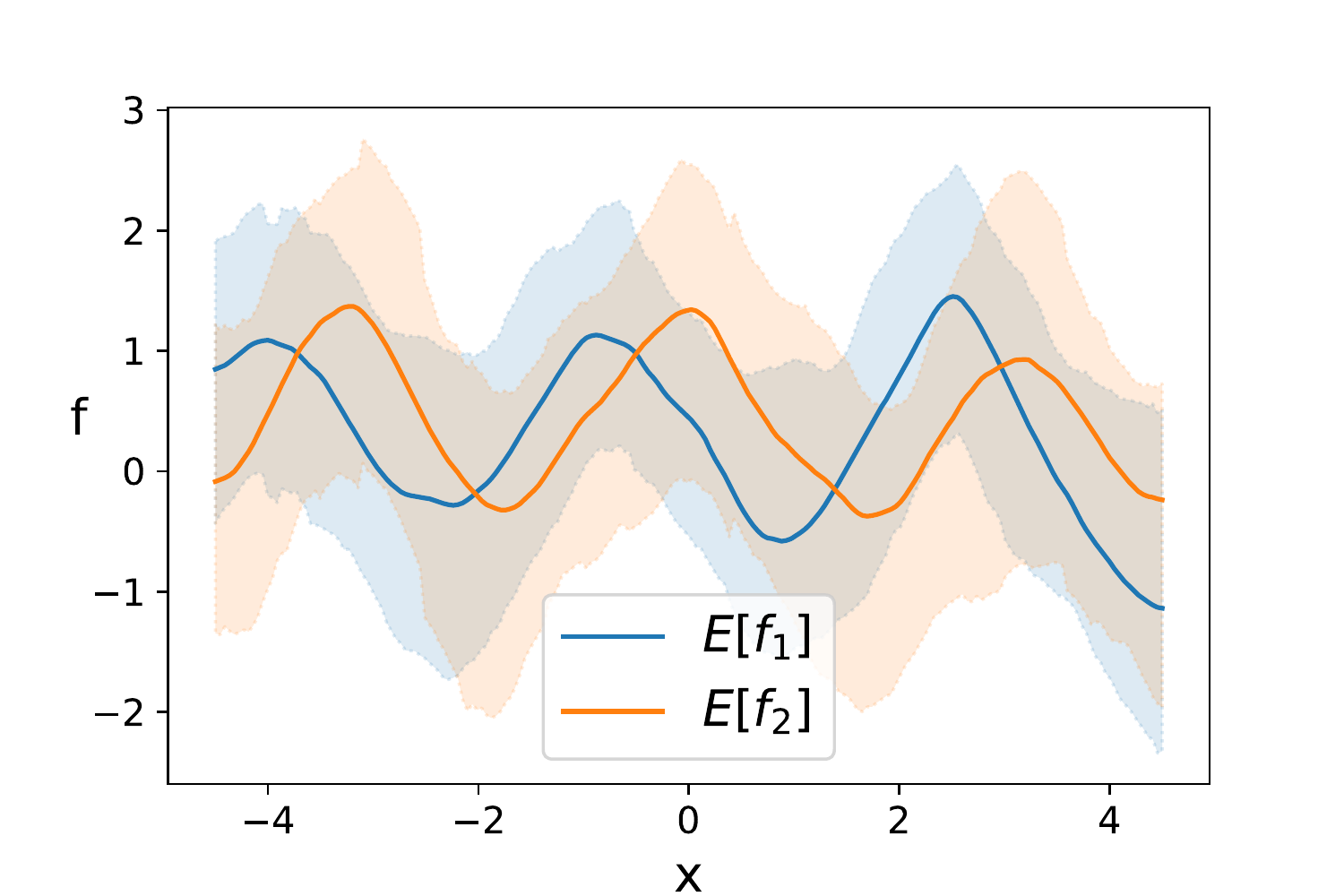} &
			\includegraphics[height=3.0cm,trim={0.8cm 0.0cm 1.5cm 0.0cm }, clip]{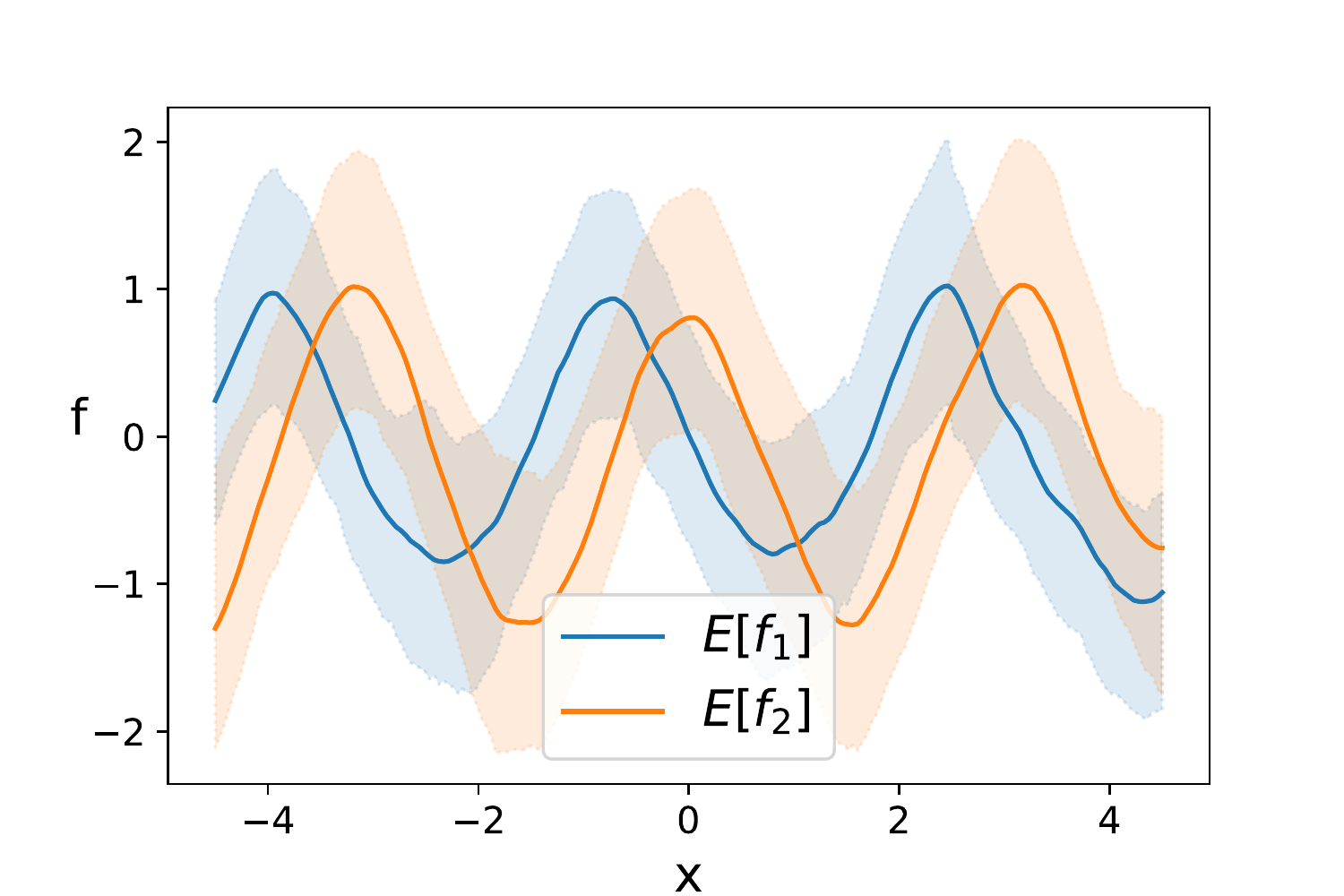} &
			\includegraphics[height=3.0cm,trim={0.8cm 0.0cm 1.5cm 0.0cm }, clip]{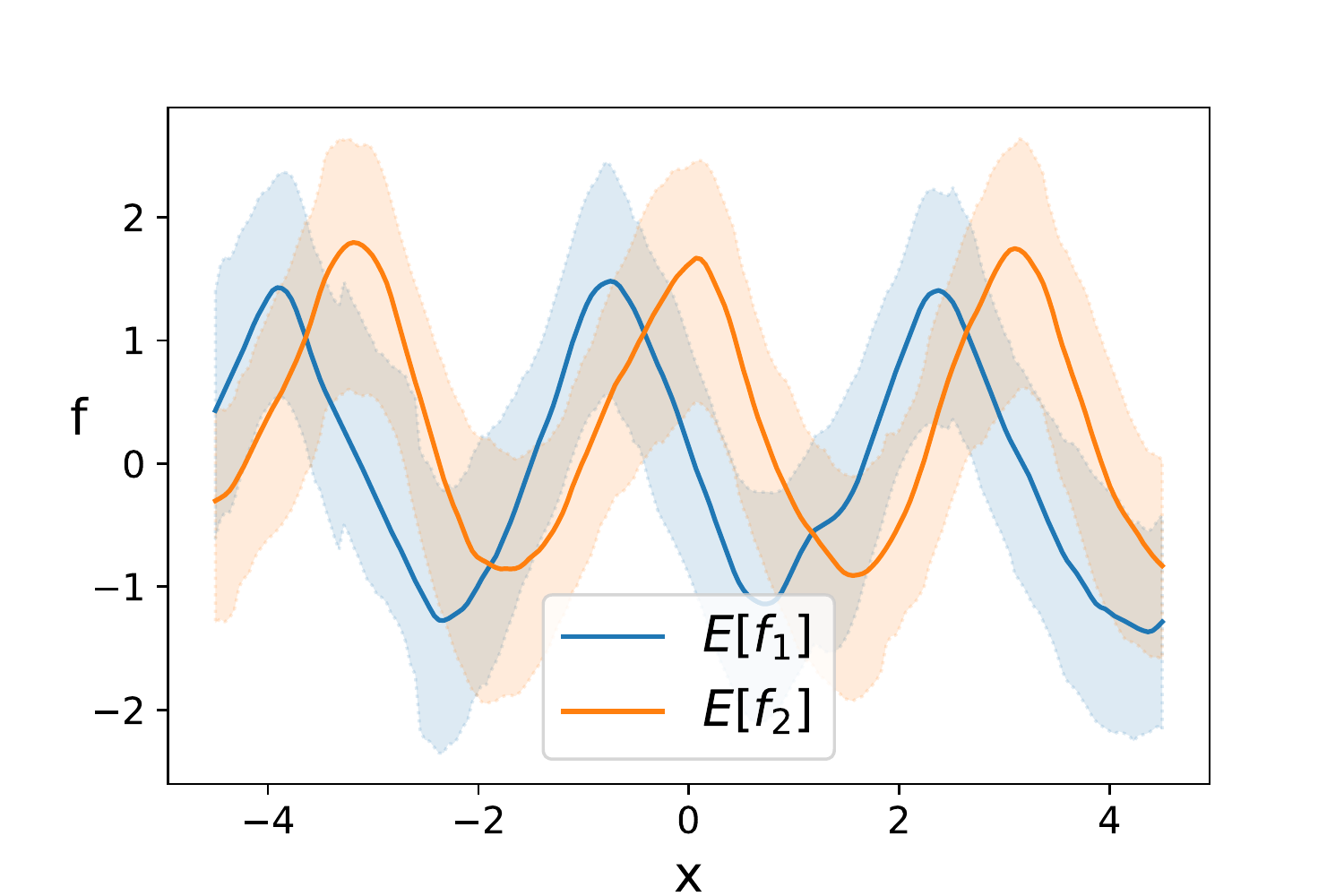} \\  
			$N=50,~|A_k|=3$ & $N=50,~|A_k|=5$ & $N=150,~|A_k|=3$ & $N=150,~|A_k|=5$\\ 
	\end{tabular}  
	\caption{Posterior mean and $95\%$ credible intervals of the two latent functions for the four artificial datasets.}
	\label{fig:post}
\end{figure*}

We use ${\bf g}$ to define a choice function. For instance, consider the set of options $A_k=\{-1,0,2\}$, given that
$$
\begin{aligned}
 {\bf g}(-1)&=[-0.416,-0.909]\\
 {\bf g}( 0)&=[1,0]\\
{\bf g}( 2)&=[-0.65,0.75]\\
\end{aligned}
$$
we have that $C(A_k)=\{0,2\}$ and $R(A_k)=A_k \backslash C(A_k)=\{-1\}$. In fact, one can notice that  $[1,0]$ dominates  $[-0.416,-0.909]$ on both the objectives, and $[1,0]$  and $[-0.65,0.75]$ are incomparable.  We sample $200$ inputs $x_i$ at random in $[-4.5,4.5]$ and, using the above approach, we generate
\begin{itemize}
 \item  $N=50$ random subsets $\{A_k\}_{k=1}^N$ of the 200 points each one of size $|A_k|=3$ (respectively $|A_k|=5$) and computed the corresponding choice pairs $(C(A_k),A_k)$ based on  ${\bf g}$;
    \item  $N=150$ random subsets $\{A_k\}_{k=1}^N$ each one of size $|A_k|=3$ (respectively $|A_k|=5$) and computed the corresponding choice pairs $(C(A_k),A_k)$ based on  ${\bf g}$;
\end{itemize}
for a total of four different datasets. Fixing the latent dimension $n_e=2$, we then compute the posterior means and $95\%$ credible intervals of the latent functions learned using the model introduced in Section \ref{sec:posterior}.
The four posterior plots are shown in Figure \ref{fig:post}. By comparing the 1st with the 3rd plot and the 2nd with the 4th plot, it can be noticed how the posterior means become more accurate (and the credible interval smaller) at the increase of the size dataset (from N=50 to N=150 choice-sets). 
By comparing the 1st with the 2nd plot and the 3rd with the 4th plot, it is evident that estimating the latent function becomes more complex at the increase of $|A_k|$. The reason is not difficult to understand. Given $A_k$, $R(A_k)$ includes the set of rejected options. These are options that are dominated by (at least) one of the options in $C(A_k)$, but we do not know which one(s). This uncertainty increases with the size of $|A_k|$, which makes the estimation problem more difficult.

Considering the same $200$ inputs $x_i$ generated at random in $[-4.5,4.5]$, we add Gaussian noise to ${\bf g}$ with $\sigma=0.1$ 
and generate two new training datasets with $|A_k|=3$ and (i) $N=100$; (ii) $N=300$. We aim to compute the predictive accuracy of the GP-based model for inferring $C(A_l)$ from the set of options $A_l$ on additional $300$ unseen pairs $\{C(A_l),A_l\}$. We denote the model learned using the $N=100$ and $N=300$ dataset respectively by Choice-GP100 and Choice-GP300. We compare their accuracy with that of an independent GP regression model which has direct access to  ${\bf g}$: that is we modelled each dimension of ${\bf g}$ as an independent GP and compute the posterior of the components of ${\bf g}$ using GP regression from $200$ data pairs $(x_i,g_1(x_i))$ and, respectively, $(x_i,g_2(x_i))$. We refer to this model as Oracle-GP: ``Oracle'' because it can directly query  ${\bf g}$. The accuracy is:
 \begin{center}\vspace{-0.2cm}
   \scalebox{0.85}{
\begin{tabular}{ccc}
    \textbf{Choice-GP100}  &   \textbf{Choice-GP300} &   \textbf{Oracle-GP}\\
\hline
 0.54 & 0.72 & 0.77\\
\end{tabular}}\vspace{-0.2cm}
 \end{center}
 averaged over $5$ repetitions of the above data generation process.  It can be noticed that the increase of $N$, the accuracy of Choice-GP gets closer to that of Oracle-GP. This confirms the goodness of the learning framework developed in this paper.

\paragraph{Real-datasets}
We now focus on four benchmark
datasets for multi-output regression problems. 
 Table \ref{tab:charac} displays the characteristics of the considered datasets. \vspace{-0.15cm}
\begin{table}[H]
		\begin{center}
			{\small
			   \scalebox{0.8}{
				\begin{tabular}{lccc}
					\hline
					{\bf Dataset} & {\bf \#Instances} & {\bf \#Attributes} & {\bf \#Outputs} \\
					\hline
					enb & 768 & 6 & 2 \\
					jura & 359 & 6 & 2 \\
                    real-estate & 414 & 5 & 2 \\
					slump & 103 & 7 & 2 \\
					\hline
				\end{tabular}}
			}
		\end{center}
		\caption{Characteristics of the datasets.}
		\label{tab:charac}
	\end{table}\vspace{-0.5cm}
More details on the used datasets are in the supplementary material. By using 5-fold cross-validation, we divide the dataset in training and testing pairs. The target values in  the training set  are used to generate  choice functions based pairs $(C(A_k),A_k)$ with $|A_k|=3$ and (i) $N=100$; (ii) $N=300$. From the test dataset, we generated $N=200$ pairs.
 As before we denote the model learned using the $N=100$ and $N=300$ dataset respectively by Choice-GP1 and Choice-GP3 and compare their accuracy  against that of Oracle-GP (learned on the training dataset by independent GP regression). The accuracy is:

 \begin{center}
   \scalebox{0.8}{
\begin{tabular}{c|cccc}
&    \textbf{Choice-GP100}  &   \textbf{Choice-GP300} &   \textbf{Oracle-GP}\\
\hline
enb  & 0.74 & 0.77&0.77\\
jura & 0.44 & 0.47 & 0.53\\
real-estate & 0.50 & 0.60 & 0.64\\
slump & 0.26 &  0.39 & 0.45
\end{tabular}}
 \end{center}
 As before, it can be noticed that the increase of $N$, the accuracy of Choice-GP gets closer to that of Oracle-GP. 

\begin{figure*}[htp!]
	\centering
	\begin{tabular}{ll}
		\includegraphics[height=3.8cm,trim={0.6cm 0.0cm 0.0cm 0.0cm }, clip]{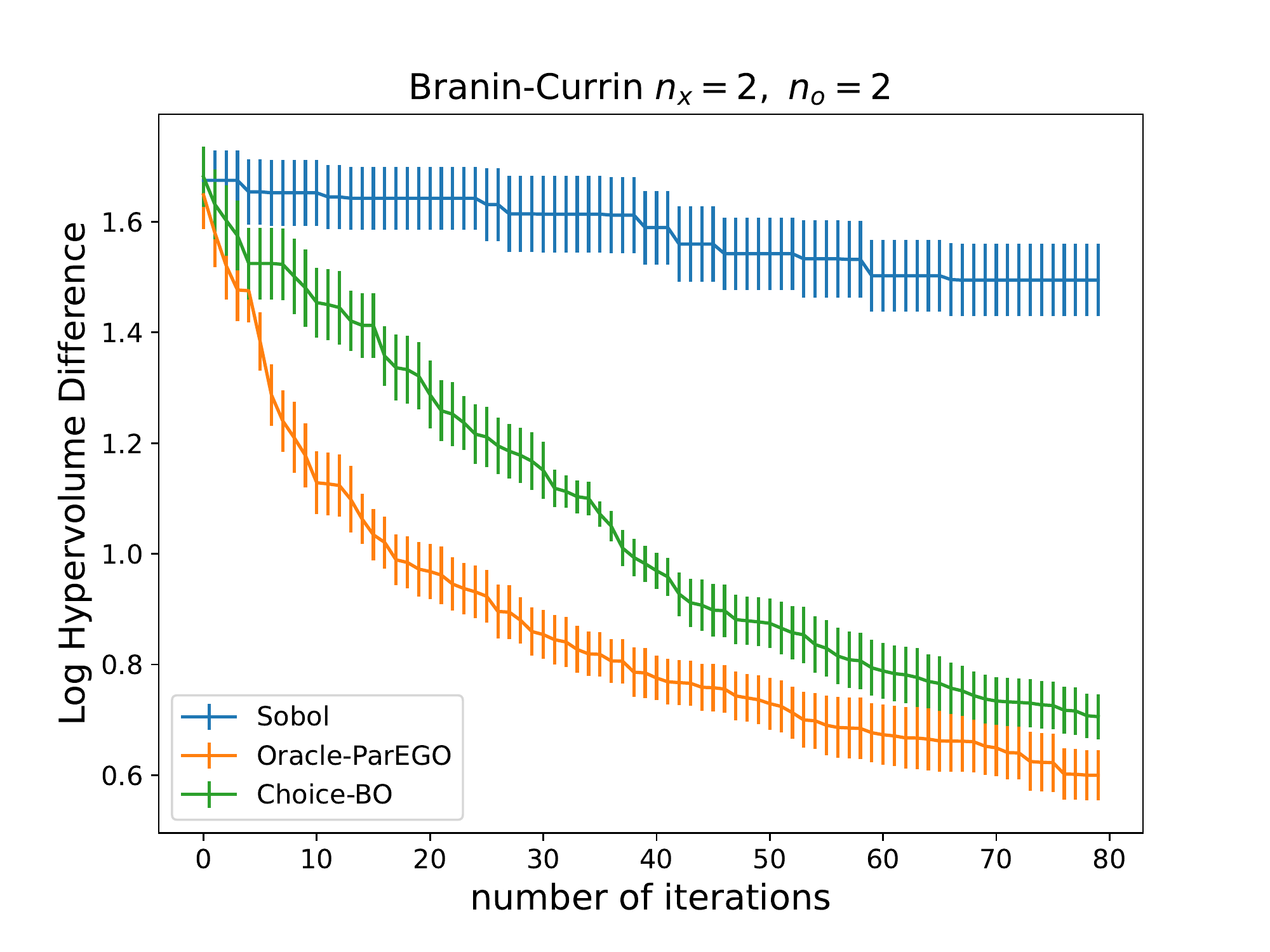}
		\includegraphics[height=3.8cm,trim={0.5cm 0.0cm 0.0cm 0.0cm }, clip]{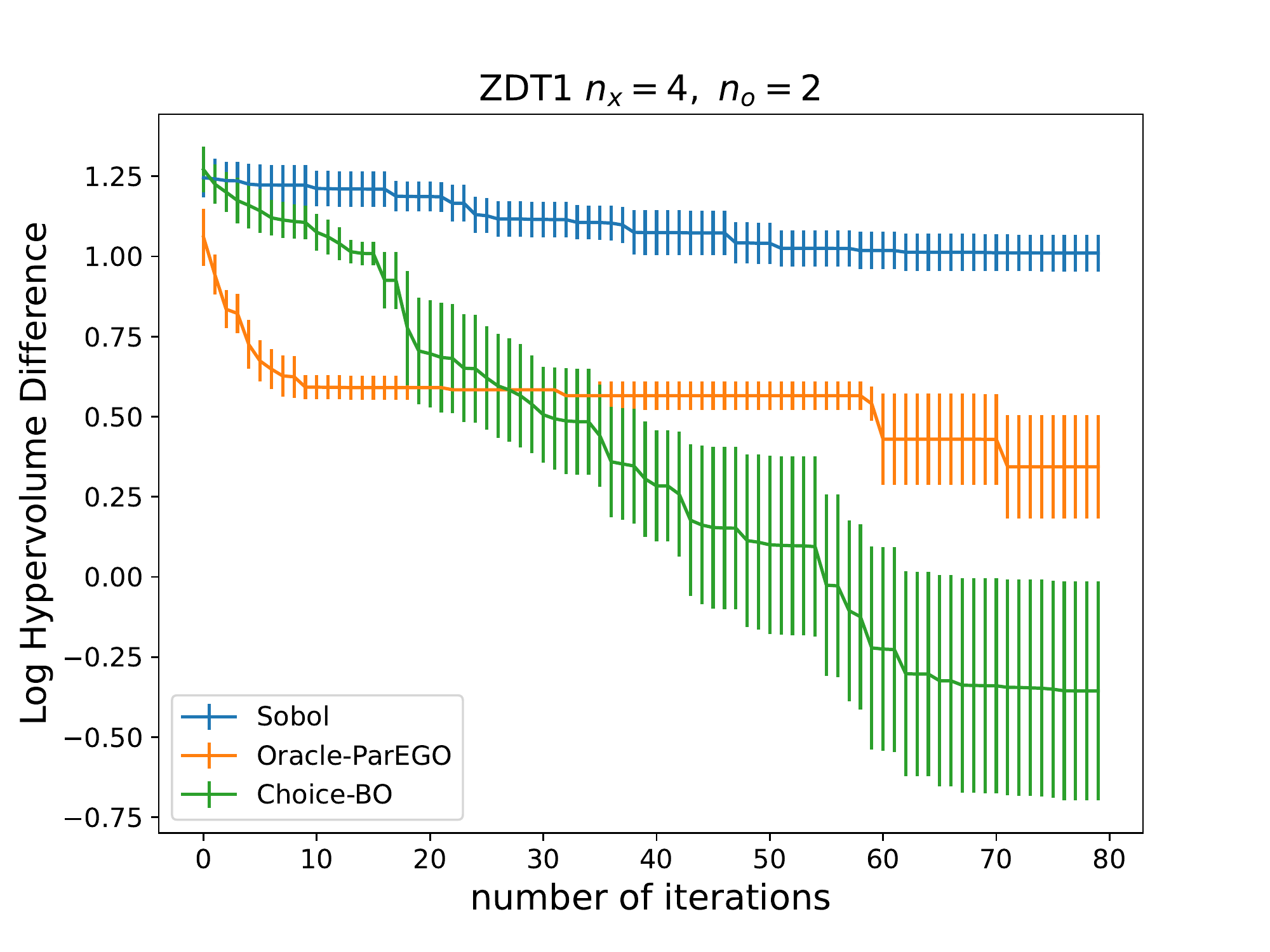}
			\includegraphics[height=3.8cm,trim={0.48cm 0.0cm 0.0cm 0.0cm }, clip]{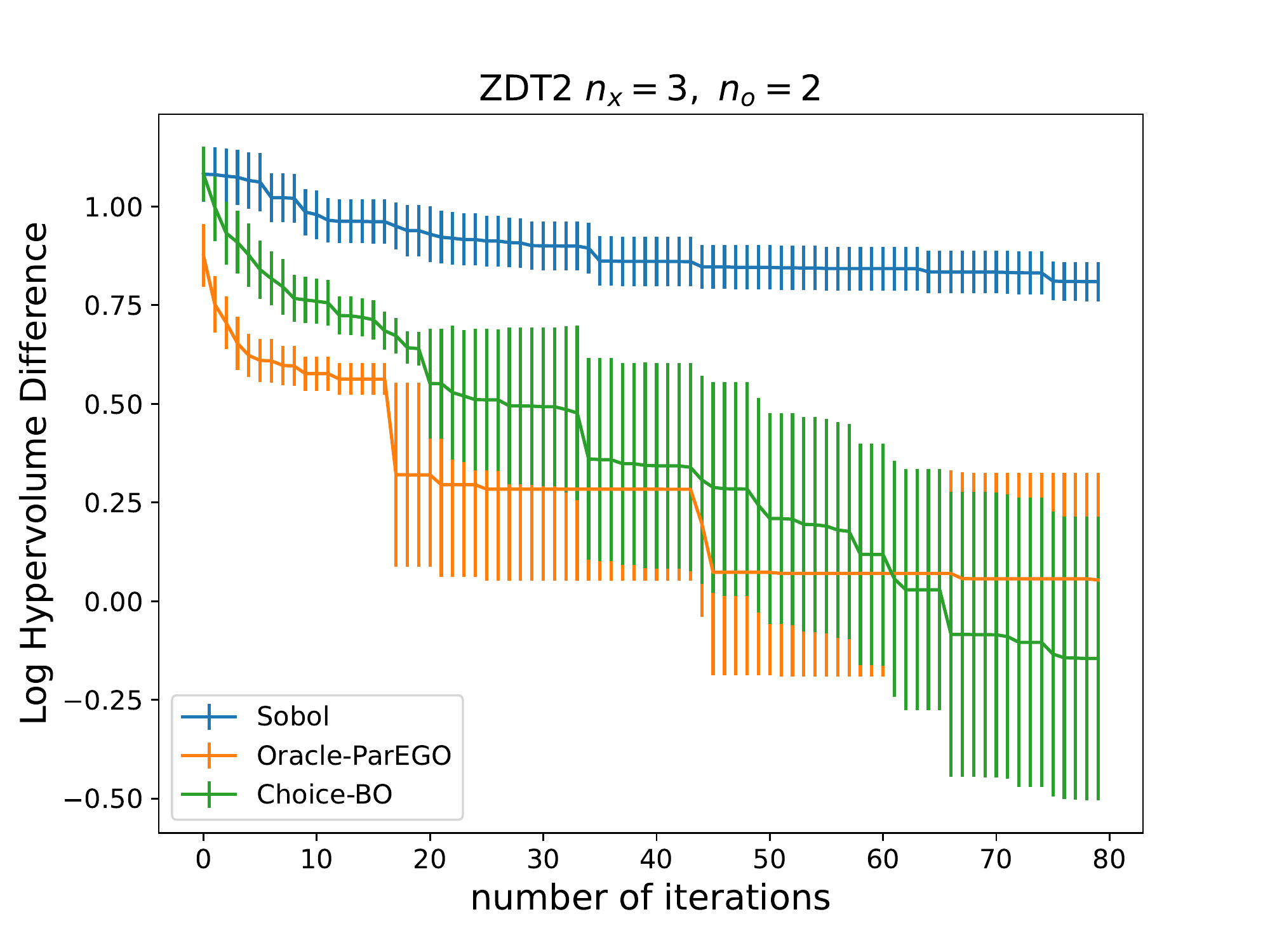}
\\
			\includegraphics[height=3.8cm,trim={0.5cm 0.0cm 0.0cm 0.0cm }, clip]{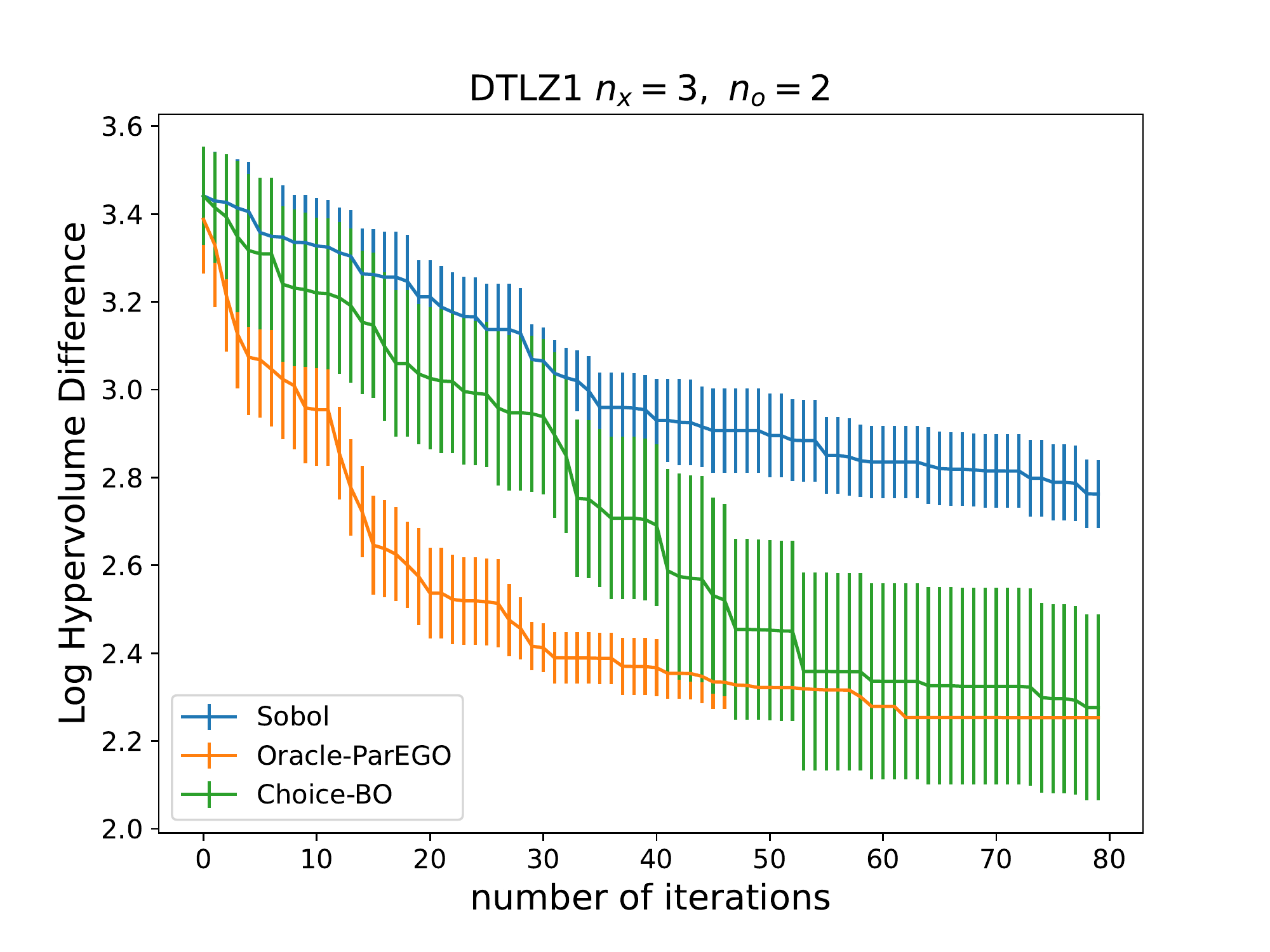}   
		\includegraphics[height=3.8cm,trim={0.5cm 0.0cm 0.0cm 0.0cm }, clip]{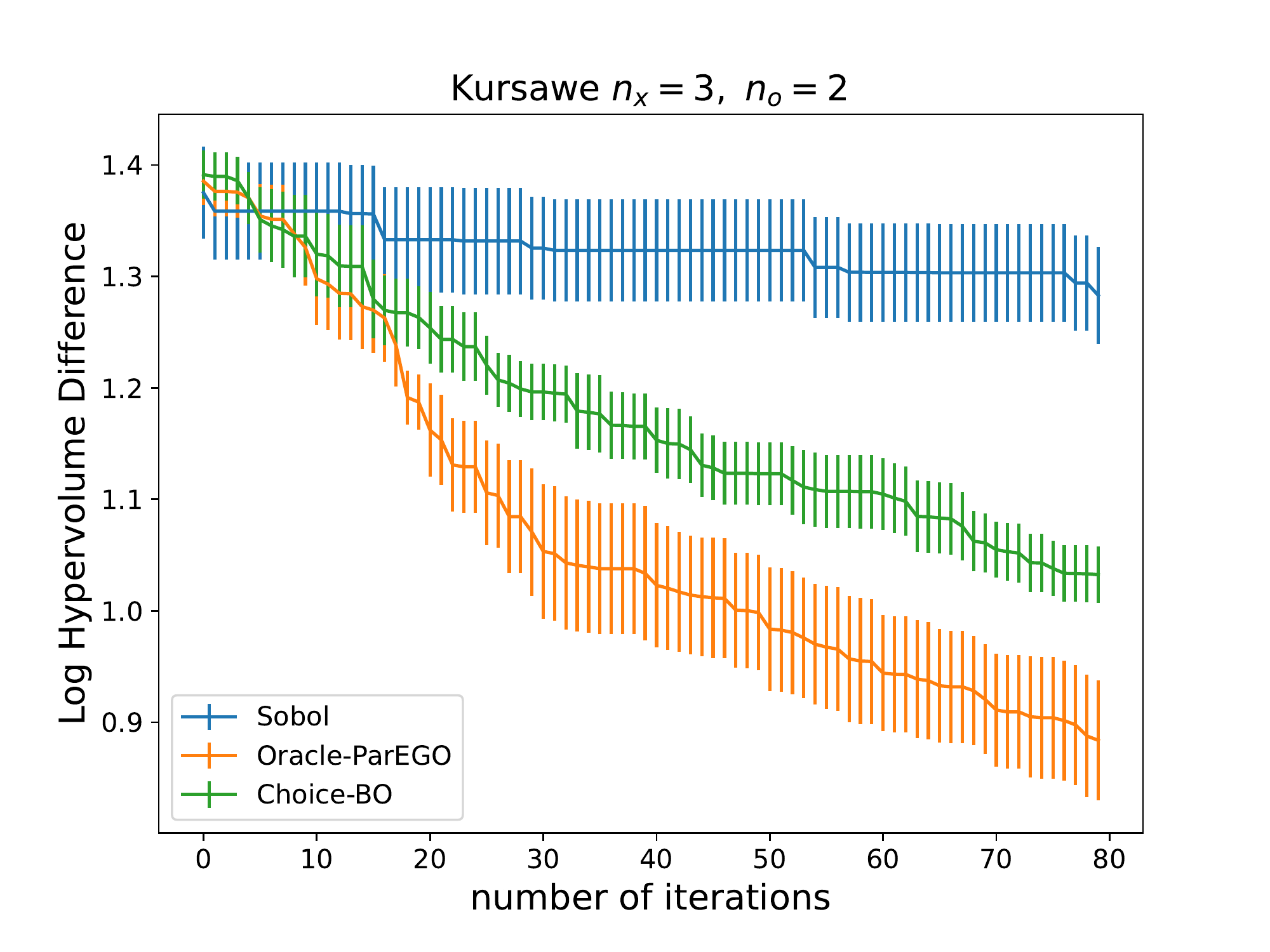}    	\includegraphics[height=3.8cm,trim={0.5cm 0.0cm 0.0cm 0.0cm }, clip]{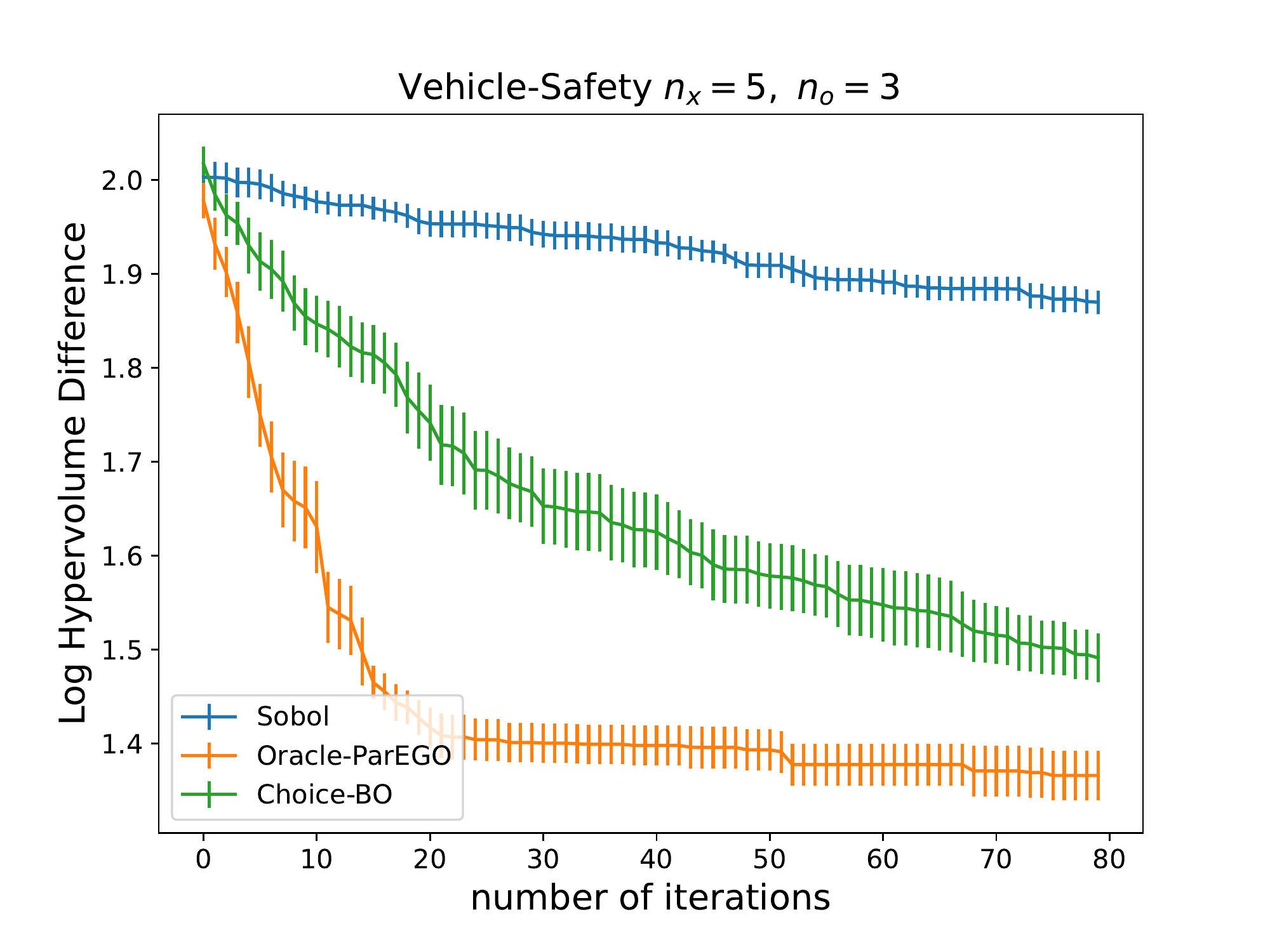}

	\end{tabular}  
	\caption{Results over 15 repetitions.  x-axis denotes  the 		number of iterations and y-axis the log-hypervolume difference.}
	\label{fig:BO}
\end{figure*} 

\subsection{Bayesian Optimisation}
We have considered for ${\bf g}({\bf x})$ five standard multi-objective benchmark functions: Branin-Currin ($n_x = 2$, $n_o = 2$), ZDT1 ($n_x = 4$, $n_o = 2$), ZDT2 ($n_x = 3$, $n_o = 2$), DTLZ1  ($n_x = 3$, $n_o = 2$), Kursawe ($n_x = 3$, $n_o = 2$) and Vehicle-Safety\footnote{The problem of determining the thickness of five reinforced components of a vehicle's frontal frame \cite{yang2005metamodeling}. This problem was previously considered as benchmark in \cite{daulton2020differentiable}.} ($n_x = 5$, $n_o = 3$). These are minimization problems, which we converted  into maximizations
	so that the acquisition function in Section \ref{sec:acq} is well-defined. We compare the Choice-GP BO (with $n_e=n_o$) approach proposed in this paper against ParEGO.\footnote{We use the BoTorch implementation \cite{balandat2020botorch}.}  
For ParEGO, we assume the algorithm can query directly  ${\bf g}({\bf x})$ and, therefore, we refer to it as Oracle-ParEGO.\footnote{The most recent MO BO approaches mentioned in Section \ref{sec:intro} outperform ParEGO. We use ParEGO only as an Oracle reference.}
Conversely, Choice-GP BO can only query ${\bf g}({\bf x})$ via choice functions. We select $|A_k|=5$ and use UCB as acquisition function for Choice-GP BO.  We also consider a quasi-random baseline that selects candidates from a Sobol sequence denoted as `Sobol`. We evaluate optimization performance on the five benchmark problems in
terms of log-hypervolume difference, which is defined as the difference between the hypervolume of
the true Pareto front\footnote{This known for the six benchmarks.} and the hypervolume of the approximate  Pareto front based on the observed data $\mathcal{X}$.
Each experiment starts with $20$ initial (randomly selected) input points which are used to initialise Oracle-ParEGO. We generate $7$ pairs $\{C(A_k),A_k\}$ of size $|A_k|=5$ by randomly selecting $7$ subsets $A_k$ of these $20$ points. These choices $\{C(A_k),A_k\}_{k=1}^7$ are used to initialise 
Choice-GP BO. A total budget of $80$ iterations are run for both the algorithms. Further, each experiment is repeated 15 times  with different initialization. In these experiments we optimize the kernel hyperparameters 
by maximising the marginal likelihood for Oracle-ParEGO  and its variational approximation for Choice-GP.
Figure \ref{fig:BO} reports the performance of the 
three methods. Focusing on Branin-Currin, DTLZ1, Kursawe, and Vehicle-Safety, it can be noticed how Choice-GP BO  convergences to the performance of the Oracle-ParEGO at the increase of the number of iterations. The convergence is clearly slower because Choice-GP BO uses qualitative data (choice functions) while Oracle-ParEGO uses quantitative data (it has directly access to ${\bf g}$). However, the overall performance shows that the proposed approach is very effective. In DLTZ1 and DLTZ2, Choice-GP BO outperforms Oracle-ParEGO. The bad performance of Oracle-ParEGO is due to the used acquisition function, which does not correctly balance exploitation-exploration in these two benchmarks. Instead, the UCB acquisition function for Choice-GP BO works well in all the benchmarks.\\
\textbf{Computational complexity:} The simulations were performed in a standard laptop. On average, the time to learn the surrogate model and optimise the acquisition function goes from 30s ($N=20$) to 180s ($N=80+20=100$).

\subsection{Unknown latent dimension}  
\label{sec:latsim}
We assume that $n_e$ is unknown and evaluate the latent dimension selection procedure proposed in Section \ref{sec:lat}. First, we consider the one-dimensional $ g(x)=\cos(2x)$ and so $n_o=1$. We generate $10$ training datasets $(C(A_k),A_k)$ with $|A_k|=3$ and sizes $N=30$ and, respectively, $N=300$ and 10 test datasets with size $300$. The following table reports the PSIS-LOO (averaged over the five repetitions) and the average accuracy on the test set for four Choice-GP models with latent dimension $n_e=1,2,3,4$.\vspace{-0.2cm} 
\begin{center}
\scalebox{0.68}{
\begin{tabular}{|l|c|c|c|c|}
 \hline
 \rowcolor{celestialblue}
 & \multicolumn{2}{c|}{\textbf{N=30}} & \multicolumn{2}{c|}{\textbf{N=300}}\\
 \hline
$n_e$ & PSIS-LOO & acc.\ test & PSIS-LOO & acc.\ test\\
 \hline
\textbf{1} (10/10) & -10 & 0.75 & -75 & 0.93\\
2 &-35 & 0.64 & -165 & 0.91\\
3 & -44 & 0.64 & -333 & 0.86\\
4 & -69& 0.62 & -388 & 0.84\\
 \hline
\end{tabular}}
\end{center}

By using PSIS-LOO (computed on the training dataset) as latent dimension selection criterion, we were able to correctly select the true latent dimension in all the repetitions (10 out of 10). The selected model has also the highest accuracy in the test set.
We now focus on the bi-dimensional vector function ${\bf g}(x)=[\cos(2x),-sin(2 x)]^\top$ and consider three different sizes for the training dataset $N=30,50,300$. \vspace{-0.3cm} 

\begin{center}
\scalebox{0.68}{
\begin{tabular}{|l|c|c|c|c|c|c|}
 \hline
  \rowcolor{celestialblue}
 & \multicolumn{2}{c|}{\textbf{N=30}} & \multicolumn{2}{|c|}{\textbf{N=50}} & \multicolumn{2}{|c|}{\textbf{N=300}}\\
 \hline
$n_e$ & PSIS-LOO & acc.\ test & PSIS-LOO & acc.\ test & PSIS-LOO & acc.\ test\\
 \hline
1 & -56 & 0.20 &-89 & 0.23 & -493 & 0.30\\
\textbf{2}  & -39 & 0.32 & -47 & 0.51 & -236 & 0.72\\
3 & -39 & 0.32& -49 & 0.49 & -269 & 0.65\\
4 & -42 & 0.30 & -53 & 0.43 & -277 & 0.64\\
 \hline
\end{tabular}}
\end{center}\vspace{-0.1cm} 

For $N=30$,  $n_e=2$ has the highest PSIS-LOO in 4/10 cases, $n_e=3$ in 4/10 cases and $n_e=4$ in 2/10 cases. 
 For $N=50$,  $n_e=2$ has the highest PSIS-LOO in 6/10 cases and  $n_e=3$ in 4/10 cases.  For $N=300$, the PSIS-LOO selects $n_e=2$ in 10/10 cases.  Note that, $n_e=1$ is never selected. Considering that the models are nested $\mathcal{M}_{1} \subset \mathcal{M}_{2} \subset \mathcal{M}_{3}...$, this shows that the selection procedure works well even with small datasets by never selecting a latent dimension that is smaller than the actual one. Moreover,  PSIS-LOO is able to select the correct dimension ($n_e=2$) at the increase of $N$.  In the supplementary material, we have reported a similar analysis for the datasets in Table \ref{tab:charac} confirming that the procedure also works for real-datasets.
 
 \section{Conclusions}
 We have developed a Bayesian method to learn choice functions from data and applied to choice function based Bayesian Optimisation (BO). As future work, we plan to develop strategies to speed up the learning process by exploring more efficient ways to express the likelihood. We also intend to explore different acquisition functions for choice function BO.

\bibliographystyle{ieeetr}

\bibliography{biblio}

\end{document}